\documentclass[letterpaper]{article} % DO NOT CHANGE THIS
\usepackage{aaai24}  % DO NOT CHANGE THIS
\usepackage{times}  % DO NOT CHANGE THIS
\usepackage{helvet}  % DO NOT CHANGE THIS
\usepackage{courier}  % DO NOT CHANGE THIS
\usepackage[hyphens]{url}  % DO NOT CHANGE THIS
\usepackage{graphicx} % DO NOT CHANGE THIS
\usepackage{natbib}  % DO NOT CHANGE THIS AND DO NOT ADD ANY OPTIONS TO IT
\usepackage{caption} % DO NOT CHANGE THIS AND DO NOT ADD ANY OPTIONS TO IT
\usepackage{booktabs}
\usepackage{makecell}
\usepackage{graphicx}
\usepackage{caption}
\usepackage{arydshln}
\usepackage{multirow}

\newcommand{\dsse}{{\textsc{AToKe-SE}}}
\newcommand{\dsme}{{\textsc{AToKe-ME}}}
\newcommand{\dsee}{{\textsc{AToKe-EE}}}
\usepackage{graphicx}
\usepackage{subcaption}
\usepackage{amssymb}
\usepackage{dsfont}

\usepackage{algorithm}
\usepackage{algorithmic}

\usepackage{newfloat}
\usepackage{listings}
\DeclareCaptionStyle{ruled}{labelfont=normalfont,labelsep=colon,strut=off} % DO NOT CHANGE THIS
\floatstyle{ruled}
\newfloat{listing}{tb}{lst}{}
\floatname{listing}{Listing}
\title{History Matters: Temporal Knowledge Editing in Large Language Model}
\author{
    Xunjian Yin\textsuperscript{\rm 1,2},
    Jin Jiang\textsuperscript{\rm 1},
    Liming Yang\textsuperscript{\rm 3},
    Xiaojun Wan\textsuperscript{\rm 1,2}
}
\affiliations{
    \textsuperscript{\rm 1}Wangxuan Institute of Computer Technology, Peking University\\
    \textsuperscript{\rm 2}Center for Data Science, Peking University\\
    \textsuperscript{\rm 3}School of Law, Tsinghua University\\
    \{xjyin, wanxiaojun\}@pku.edu.cn, jiangjin@stu.pku.edu.cn, yanglm23@mails.tsinghua.edu.cn
}

\usepackage{bibentry}
\begin{document}

\maketitle

\begin{abstract}
The imperative task of revising or updating the knowledge stored within large language models arises from two distinct sources: intrinsic errors inherent in the model which should be corrected and outdated knowledge due to external shifts in the real world which should be updated.
Prevailing efforts in model editing conflate these two distinct categories of edits arising from distinct reasons and directly modify the original knowledge in models into new knowledge.
However, we argue that preserving the model's original knowledge remains pertinent. Specifically, if a model's knowledge becomes outdated due to evolving worldly dynamics, it should retain recollection of the historical knowledge while integrating the newfound knowledge.
% If we update Leo Messi's team as Inter Miami, then Messi's \textit{last team} should naturally be Paris Saint-Germain.
In this work, we introduce the task of \textbf{Temporal Knowledge Editing} (\textbf{TKE}) and establish a benchmark \textsc{AToKe} (\textbf{A}ssessment of \textbf{T}emp\textbf{O}ral  \textbf{K}nowledge \textbf{E}diting) to evaluate current model editing methods.
We find that while existing model editing methods are effective at making models remember new knowledge, the edited model catastrophically forgets historical knowledge.
To address this gap, we propose a simple and general framework termed \textbf{M}ulti-\textbf{E}diting with \textbf{T}ime \textbf{O}bjective (\textbf{METO}) for enhancing existing editing models, which edits both historical and new knowledge concurrently and optimizes the model's prediction for the time of each fact.
Our assessments demonstrate that while \textsc{AToKe} is still difficult, METO maintains the effectiveness of learning new knowledge and meanwhile substantially improves the performance of edited models on utilizing historical knowledge.
\end{abstract}

\section{Introduction}
Large-scale language models (LLMs) have made impressive progress in the last few years \citep{brown2020language,ouyang2022training,touvron2023llama,openai2023gpt4}. However, the knowledge in a language model always needs to be updated because the internal knowledge of the model can be wrong and the external world knowledge is constantly changing \cite{DBLP:conf/iclr/SinitsinPPPB20,hartvigsen-etal-2022-toxigen,10.1145/3571730}. It would be costly to retrain the model each time, so there is a lot of work proposing knowledge editing (KE) methods that allow new correct knowledge to be injected directly into specific model parameters.
Previous work includes constrained fine-tuning~\cite{zhu2020modifying}, hypernetwork knowledge editing~\cite{DBLP:conf/emnlp/CaoAT21,hase2021language,DBLP:conf/iclr/MitchellLBFM22}, external memory-based editing~\cite{DBLP:conf/icml/MitchellLBMF22,zhong2023mquake,zheng2023edit} and locate-then-edit model editing~\cite{dai-etal-2022-knowledge,meng2022massediting}. All of these methodologies focus on making the model memorize new knowledge.

\begin{figure}[]  
    \centering  
    \includegraphics[width=0.45\textwidth]{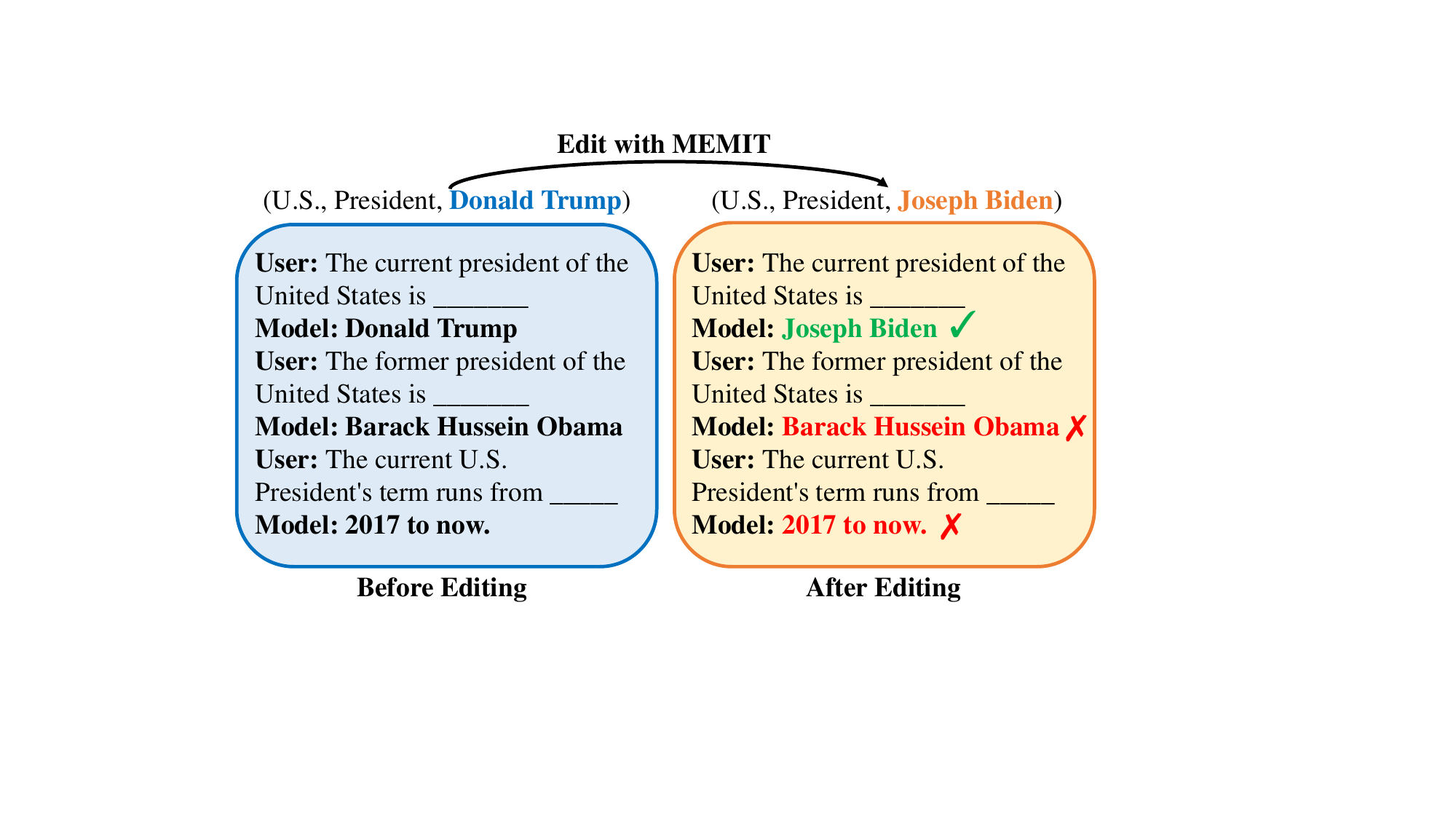} 
    \caption{An example of an edited GPT-J model losing historical knowledge by using the existing editing method MEMIT. The editing operation overwrites \textit{Joseph Biden being the President of the U.S.} against \textit{Donald Trump}, but does not preserve historical knowledge, which causes Trump's relationship with the U.S. President to be lost.}  
    \label{fig: unedited}  
\end{figure}

However, a clear differentiation is essential during the process of knowledge editing, specifically discerning between two distinct scenarios: (1) \textbf{knowledge correction}, involving rectification of inaccurate knowledge arising from the model's training data which needs to be corrected, and (2) \textbf{knowledge updating}, necessitated by shifts in the external world or evolving cognitive paradigms which needs to be updated.
Existing work on KE lacks this crucial distinction, and is committed to making the LLMs memorize the new knowledge, while simply ignoring the original knowledge in LLMs.
However, within the context of knowledge updating, we believe that retaining \textbf{historical knowledge} within the model holds great value.
For example, as shown in Figure \ref{fig: unedited}, although we are all ``updated" with the knowledge that \textit{President of the United States is Joseph Biden}, we still sometimes want to know who \textit{the former President} was.

% For example, if we update that \textit{the team Leo Messi plays for is Inter Miami}, then the \textit{last team} Messi played for was naturally Paris Saint-Germain (PSG), which shouldn't be forgotten either.

\begin{figure*}[htbp]  
    \centering  
    \includegraphics[width=0.95\textwidth]{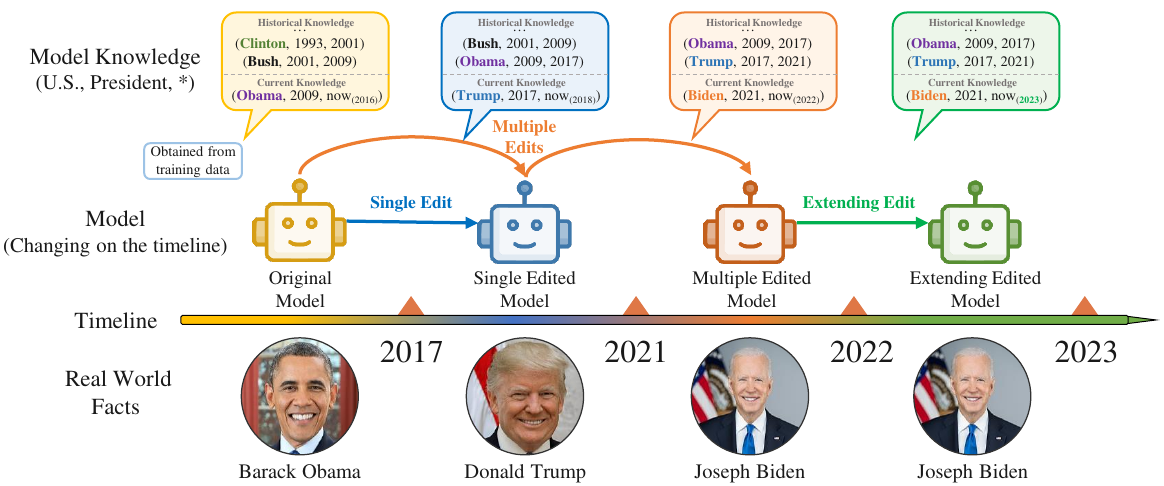} 
    \caption{Presentation of Temporal Knowledge Editing Task. Suppose that the training corpus for the model is collected in 2016, so the internal time of the model is 2016. After one edit the model can obtain the knowledge that \textit{Donald Trump is the president of the United States in 2017 and Obama is the former president}. Then keep editing in this way, the model's internal time keeps moving forward while not losing history and ensuring temporal ordering.}  
    \label{fig:task}  
\end{figure*}

The current evaluation of KE methods focuses on new knowledge being memorized and irrelevant knowledge being left unaltered, while not considering the appropriate keeping of relevant historical knowledge.
Therefore, we propose the task of Temporal Knowledge Editing (TKE) and construct a benchmark \textsc{\textbf{AToKe}} for evaluating KE methods in historical knowledge preservation, by collecting series of world knowledge with timestamps and viewing them as the form of a series of knowledge updates.
We explore single editing, multiple editing and extending editing (e.g., letting the model know that \textit{the president is still Biden in 2023} in Figure \ref{fig:task}) of facts on the time series. 
After each editing, we ask questions about historical knowledge and current knowledge. 
%In addition, to be more realistic, the format of the question includes both \textit{explicit time question}, which refers to asking the question that have an explicit time frame and \textit{relative time question}, which does not contain an explicit time frame and uses words like ``previous" or ``last one" instead. //后面重复定义了

We evaluate  state-of-the-art knowledge-editing methods on \textsc{AToKe} and find that they are effective in making the model memorize new knowledge but cause confusion in time, e.g., in our example,  the knowledge about the former president disappeared in edited LLMs.

% To address this problem, inspired by previous work~\cite{DBLP:conf/icml/MitchellLBMF22,zhong2023mquake}, we propose a simple memory-based approach \textsc{MeTar}. \textsc{MeTar} stores each fact, both edited and original, by assigning a timestamp to it, and prompts the LLM to temporally localize it over the time series. LLMs with an edited temporal memory can perform inference and application based on the retrieval of facts from memory and recall of internal knowledge.
% \textsc{MeTar} does not affect the capabilities of the LLM itself and does not require further training or additional parameters, making it convenient and easily scalable to larger models.
% Our experiments show that \textsc{MeTar} outperforms previous methods on our benchmark \textsc{AToKe}.

In order to address this problem to some extent and to enhance the effectiveness of current model editing methods, we propose a simple and general framework, METO, which edits both historical and new knowledge and optimizes the model's prediction for the time of each fact.

Our contributions are listed below:
\begin{enumerate}
    \item We categorize knowledge editing scenarios, point out the issue that current methods corrupt historical knowledge, and validate it through experiments and analysis.
    \item We are the first to propose the Temporal Knowledge Editing task, and publish a benchmark \textsc{AToKe} for evaluating the task.
    \item We propose a new editing framework METO that improves the performance of previous editing methods on the benchmark.
\end{enumerate}

Our benchmark has been released to the community to facilitate future research \footnote{\url{https://github.com/Arvid-pku/ATOKE}}.

\section{Temporal Knowledge Editing}
This section introduces our arguments about knowledge editing as well as definitions and evaluation for the temporal knowledge editing task.
\subsection{Background: Knowledge in Language Models}
\label{background}
There has been a lot of work considering LLMs as knowledge bases and accessing, applying and manipulating the knowledge in them~\cite{petroni-etal-2019-language,alkhamissi2022review}.
Previous work always represents knowledge as a triple $(s, r, o)$, consisting of a subject ($s$), a relation ($r$), and an object ($o$) (e.g. (\textit{U.S.}, \textit{president}, \textit{Joseph Biden})) and explores how to extract knowledge from LLMs~\cite{bordes2013translating,Lin_Liu_Sun_Liu_Zhu_2015}.
In this paper, we follow the work that uses discrete prompt to extract knowledge~\cite{petroni2019language,davison-etal-2019-commonsense}.
Specifically, we construct a natural language template $q_r(\cdot)$ for each relation $r$, which is then combined with a subject $s$ in a knowledge triple to generate a question or a cloze-style statement $q_r(s)$, which is also consistent with the previous work of the knowledge editing~\cite{meng2022locating,zhong2023mquake}.
It is a natural way to extract or test the knowledge in a language model.

\subsection{Requirements for Knowledge Editing}
\label{requirements}
What are the requirements that a good knowledge editing method needs to satisfy?
We believe that the edited model should be changed in some ways and hold constant in others.
\subsubsection{Aspects of Change}
The knowledge we edit is naturally expected to change in the model, which is generally referred to as \textit{Edit Success} in prior work in evaluation. 
In addition, the paraphrase expression about the knowledge should also be changed, which is referred to as \textit{Paraphrase Success} in evaluation. 
Similarly, the knowledge associated with the edited knowledge should also be changed, which has been explored by \citet{zhong2023mquake} and referred to as \textit{Multi-hop Accuracy}.
\subsubsection{Aspects of Constancy}
Except for what is relevant to the edited knowledge, it seems like no other knowledge or ability of the model should be changed. 
Previous work~\cite{meng2022massediting} has defined \textit{Neighborhood Success} to evaluate whether other knowledge of the original entity is affected. 
They also define \textit{Reference Score} to check if generations from the edited model are semantically consistent with the new object.
And \textit{Generation Entropy} is designed to test ability of the edited model to generate fluent sentences.

However, no work has yet explored the knowledge to be edited.
We argue that if the knowledge within the model is outdated (changed to historical knowledge) due to the flow of time, then the historical knowledge should also be preserved within the model.
It is quite reasonable because if not then the timeline in the model would be messed up and our \textit{former president} would disappear in the aforementioned example.
Therefore, we propose the task of Temporal Knowledge Editing~(TKE)\footnote{Temporal Knowledge Editing focuses on knowledge that is outdated due to the flow of time and therefore needs to be edited, rather than original errors within the model.}.

\subsection{Task Definition}
As mentioned in Section \ref{background}, we use triples to represent knowledge. 
In addition, we attach temporal scope to each relational fact $\{(s,r,o,t_s,t_u)\}$ since the relational fact often shows temporal dynamics, where $t_s$ and $tu$ denote the fact $(s, r, o)$ valid from $ts$ to $tu$.
As time passes, the same subject and relation will correspond to different objects in sequence, such as (\textit{U.S.}, \textit{president}, \textit{Donald Trump}, \textit{2017}, \textit{2021}), (\textit{U.S.}, \textit{president}, \textit{Joseph Biden}, \textit{2021}, N/A) and so on.
Consistent with previous work, we use $(s,r,o,t_s,t_u) \rightarrow (s,r,o^\star,t_s^\star,t_u^\star)$ to represent an knowledge editing operation, denoted as $e$, meaning that the object of subject $s$ under relation $r$ changes from $o$ to $o^\star$.
And the reason for the edit is that $(s,r,o)$ is valid from $t_s$ until $t_u$, and after that $(s,r,o^\star)$ is valid from $t_s^\star$ to $t_u^\star$\footnote{Note that the $t_u$ is vacant if the knowledge is still valid currently.}.
Specifically, $o$ and $o^\star$ can be the same object, at which point the editing operation $(s,r,o,t_s,t_u) \rightarrow (s,r,o,t_s,t_u^\star)$ will make the model know that the fact $(s,r,o)$ lasts until $t_u^\star$, where $t_u^\star$ is later than $t_u$.
For convenience of expression, we refer to $(s,r,o,t_s,t_u)$ as \textbf{historical knowledge} and $(s,r,o^\star,t_s^\star,t_u^\star)$ as \textbf{current knowledge}.
% This operation can also be used to update the time in model cognition.

Given a collection of knowledge editing operations $\mathcal{E}=\{e_1,e_2,\dots\}$ and a language model $\mathcal{M}$, knowledge editor $F$ aims to get an edited model $\mathcal{M}_e = F(\mathcal{M}, \mathcal{E})$ that satisfy the requirements mentioned in Section \ref{requirements}.
Specifically, in temporal knowledge editing, for the edited model $\mathcal{M}$, we expect that it should not only memorize the current knowledge $(s, r, o^\star)$, but the knowledge in it should also be organized on the timeline. In other words, the model can answer correctly when asked questions about the ``previous" or ``last one" knowledge $(s, r, o)$, or when asked directly about the relevant knowledge $(s, r, ?)$ at a certain moment in time between $t_s$ and $t_u$.
For example, after updating the model's knowledge of U.S. president to Biden, the test questions include ``\textit{Who was the former president of the United States?}" or ``\textit{Who was the president of the United States in 2005?}" .

\subsection{Evaluation of Temporal Knowledge Editing}
Since time is constantly moving forward, a piece of knowledge may be updated multiple times in different temporal scopes. Therefore, we define three evaluation subtasks: 1) \textbf{Temporal Knowledge Single Editing (TSE)}, which requires that, after a single update to a piece of knowledge, the model sustains temporal chronology. 2)  \textbf{Temporal Knowledge Multiple Editing (TME)}, demanding ordered temporal succession after multiple unidirectional knowledge updates.
It is worth pointing out the difference between TME and mass-editing of prior work~\cite{meng2022massediting,DBLP:conf/iclr/MitchellLBFM22}: mass-editing refers to the simultaneous editing of multiple facts, which are usually unrelated, while TME refers to the continuous editing of the same $s$ and $r$ at different time scopes.
3) \textbf{Temporal Knowledge Extending Editing (TEE)}, which requires that duration of target knowledge perceived by the model is successfully prolonged.
More details on the benchmark are presented in Section \ref{sec: benchmark}.

\section{\textsc{AToKe}: \textbf{A}ssessment of \textbf{T}emporal \textbf{K}nowledge \textbf{E}diting}
\label{sec: benchmark}
We introduce the \textsc{AToKe} (Assessment of temporal knowledge editing) benchmark, which comprises three distinct datasets as outlined in Table \ref{tab:dataset}. These datasets are specifically designed to evaluate the effectiveness of knowledge editing methods in handling temporal information.
The first dataset, referred to as \textsc{AToKe}-SE, consists of temporal knowledge \textbf{S}ingle  \textbf{E}dits. The second dataset, \textsc{AToKe}-ME, includes temporal knowledge \textbf{M}ultiple \textbf{E}dits associated with the same subject and relation. The third dataset, \textsc{AToKe}-EE, encompasses knowledge \textbf{E}dits that \textbf{E}xtend the temporal scope of the original knowledge.
% These datasets have been meticulously curated to facilitate the assessment of knowledge editing techniques, with a specific emphasis on temporal aspects. 
In the following sections, we first present the collection of the basic temporal knowledge and further describe how the three datasets were extracted from it. We then detail the data statistics and evaluation settings of the datasets, and finally present the evaluation metrics.

\begin{table}[t]
    \centering
    \resizebox{\columnwidth}{!}{
    \begin{tabular}{cl}
    \toprule
    BTK & (United States, head of government, \\
    &(Obama, 2009, 2017), (Trump, 2017, 2021), \\
    &(Biden, 2021, 2022))\\
    \midrule
    \textbf{SE} & (United States, head of government, \\
    &(Obama, 2009, 2017) → (Trump, 2017, 2021))\\
    \addlinespace[0.1ex] % 添加空间
    \cdashline{2-2}[3pt/3pt] % 在第二列创建虚线并添加空间
    \addlinespace[0.2ex] % 添加空间
    \textbf{ME} & (United States, head of government, \\
    &(Obama, 2009, 2017) → (Trump, 2017, 2021) \\
    &→ (Biden, 2021, 2022)) \\
        \addlinespace[0.1ex] % 添加空间
    \cdashline{2-2}[3pt/3pt] % 在第二列创建虚线并添加空间
        \addlinespace[0.2ex] % 添加空间
    
    \textbf{EE} & (United States, head of government, \\
    &(Biden, 2021, 2022) → (Biden, 2021, 2023))\\
    \midrule
$\mathcal{Q}$ &a) Who is the current President of the United States? (ALL) \\
    &b) Who was the President of the United States \\
    &in the previous term? (SE\&ME) \\
    &c) Who holds the position of President in the United States \\
    &from 2017 to 2021? (SE\&ME)  \\
    &d) Who was the President of the United States \\
    &from 2009 to 2017? (SE\&ME)  \\
    &e) From 2022 to 2023, who serves as the president \\
    &of the United States? (EE) \\
    \midrule
    $\mathcal{A}$ & a) Donald Trump \\
    & b) Barack Obama \\
    & c) Donald Trump \\
    & d) Barack Obama \\
    & e) Joseph Biden \\
    \bottomrule
    \end{tabular}
    }
    \caption{An instance illustrating the construction of the dataset, where the collected base temporal knowledge (BTK) serves as the foundation. The dataset encompasses three types of edits, namely single edit (\textbf{SE}), multiple edits (\textbf{ME}), and extending edit (\textbf{EE}), along with corresponding questions ($Q$) and their respective answers ($A$) following the knowledge editing.  In this case, the model has been edited from Obama to Trump.}
    \label{tab:dataset}
\end{table}

\subsection{Basic Temporal Knowledge Collection}

\subsubsection{Sampling time series facts}

% 附录：从since-until分别抽取，然后merge，得到所有的事件的基础格式。

Our dataset is based on YAGO3.0.3\footnote{\url{https://yago-knowledge.org/downloads/yago-3}} \cite{Mahdisoltani2015YAGO3AK}, a knowledge base comprising fact triples associated with millions of entities extracted from Wikipedia, enriched with WordNet, GeoNames, and other data sources.
Following the previous work \cite{dasgupta-etal-2018-hyte}, we begin by sampling temporal facts from the YAGO data. We extract all fact triples with their time scope and obtain the temporal facts set $\{(s,r,o,t_s,t_u)\}$.
Subsequently, after matching by $s$ and $r$ and sorting by $t_s$ with all these facts, we obtain the chain of temporal facts $\mathcal{C}=\{(s,r,o_1,t_{s_1},t_{u_1}),...,(s,r,o_N,t_{s_N},t_{u_N})\}$. Finally, to ensure non-overlapping and non-contradictory time sequences within the chain, we employ a heuristic algorithm. 
% See the Appendix \ref{Appendix:Sampling} for more technical details.

\subsubsection{Locating Time of Model Facts}
\label{sec: modeltime}

For a chain of temporal facts $\{(s,r,o_N,t_{s_N},t_{u_N})\}$, its $o$, $t_s$ and $t_u$ are constantly changing. Therefore, to test the performance of knowledge editing, we need to determine where the knowledge is located in the model. First, we use the GPT-J model\footnote{Without loss of generality, we use GPT-J as the representative LLM used in the experiments.} to verify and filter out $(s,r,*)$ that the model does not have. Next, we locate the model's knowledge position in the temporal fact chain. By asking questions about the facts in the temporal chain, we find the most recent fact in each chain that the model has successfully recalled as the model's initial knowledge. 
% See Appendix \ref{Appendix:Locating} for more details of locating.

\subsubsection{Sampling Future Fake facts}
% 构造反事实的事件 
So far, we have collected a temporal chain of facts from the YAGO dataset where the latest knowledge is collected in 2022. 
To ensure that there is always new knowledge used to update for all models in our dataset and to make our dataset effective to be used as the benchmark for TKE in the long run, we design future fake facts to augment the original temporal chain. Specifically, for each chain, we sample one counterfactual object in YAGO to serve as counterfactual object $o_{N+1}$ that is in the same class as real objects.
Then $o_{N+1}$ will be randomly assigned a reasonable time scope and appended to the end of the chain. Finally, we obtain a temporal chain of facts that incorporates constructed fake facts.

\subsubsection{Generating Temporal natural language questions}
\label{sec: genq}
% 使用 chatgpt构造提问的问题
% 给定我们构造的时序事实链，我们对每一个关系r构建问题，问题的标准答案是时序链中的宾语o。构造的问题从问题的时间分为三个方面：当前问题，过去问题，时间问题。问题格式分为“自然语言问题”和“completion”两种格式。我们首先利用ChatGPT（gpt-3.5-turbo）对13个关系自动生成问题，然后进行人工筛选。我们提示ChatGPT为当前问题和过去问题生成两个问题，对时间问题生成四个问题（包含两个现在式和两个过去式问题）我们使用的prompt展示在附录3，我们构建的问题在GitHub。

% 在这里说明，生成的两个问题是paraphrase的
As mentioned in Section \ref{background},
given a chain of temporal facts that we construct, we construct template $q_r(\cdot)$ for each relation $r$. The constructed questions are categorized into two categories based on time:
1) \textbf{Explicit time question}, which refers to asking the question that have an explicit time frame. 
2) \textbf{Relative time question}, on the other hand, does not contain an explicit time frame and uses words like ``present",  ``previous" or ``last one" instead.
We first utilize ChatGPT (gpt-3.5-turbo) to automatically generate questions for all relations, and then manually filter them. 
%We prompt ChatGPT to generate two questions for current and past questions respectively, and four questions for time-related questions (including two present tense and two past tense questions). 
% The details of question generation such as prompts used, question examples and statistics of questions are shown in Appendix \ref{Appendix:Generating}.

\subsection{Construction of Three Datasets}
% 在附录中给出一个三个数据集的分别的例子。
% 吧这几个部分，单独的拿出来放三个表格，
Up until now, we have obtained the basic temporal knowledge, including temporal fact chains, questions and answers at different time points (respectively BTK, Q and A in Table \ref{tab:dataset}).  And after locating and expanding, the first fact of every temporal fact chain represents the current knowledge of the model. 
Therefore, all of our edits begin with the first fact.
Moving forward, we will construct three temporal knowledge editing datasets.
%at various levels using the temporal chains, questions, and answers.

\subsubsection{\textsc{AToKe}-SE}
% 对于单次编辑(SE)，我们遍历所有的时序事实链，从每个链中抽取两个事实构建一次编辑操作，并对模型编辑前后的事实进行提问。如表所示，对于美国的领导人这一事实链，模型当前已知美国领导人为奥巴马。我们抽取奥巴马，2009-2017到特朗普2017-2021这两个事实来构建一次事实编辑。模型编辑后，模型关于美国总统的知识转变为特朗普。针对单次编辑，我们对模型进行当前和过去的提问来评估编辑效果，同时提问带有时间范围的问题。

For Single Edit (SE), we take the first two facts from each chain in BTK to construct a single edit operation, and ask questions about both historical and current facts. 
%As shown in Table \ref{tab:dataset}, for the fact chain of leaders of the United States, the model currently knows that the leader of the United States is Barack Obama. We extract two facts from Obama, 2009-2017 to Trump 2017-2021 to construct one fact edit. After the model edit, the model's knowledge about the U.S. president shifts to Trump. For the single edit, we ask current and past questions to the model to assess the effect of the edit, as well as questions with a time range (From ... to ...).

\subsubsection{\textsc{AToKe}-ME}
% 多次编辑（ME）是单次编辑的叠加。我们同样遍历所有时序链，对每条链中所有事实构建连续的编辑操作，并对每个时间范围的事实进行提问。在表中，美国领导人会经过从奥巴马到特朗普，特朗普到拜登两次编辑。特殊地，我们会在每次编辑后对模型进行“现在”和“过去”的提问。在所有编辑完成后，对模型进行带有时间范围的提问。

Multiple edits (ME) are superimposed on single edits. We similarly traverse all temporal chains, construct successive editing operations for all facts in each chain, and ask questions about facts with each time scope. 
% In Table \ref{tab:dataset}, U.S. leaders go through two edits from Obama to Trump and Trump to Biden. Specifically, we ask the model questions about the ``present" and ``past" after each edit. After all edits have been made, the model is asked a question with a time range.

\subsubsection{\textsc{AToKe}-EE}
% 扩展编辑（EE）是对模型当前事实的时间范围进行扩展的编辑操作。我们遍历所有的时序事实链，并选择每条链中的第一个事实作为扩展对象。通过人工的方式对该对象进行扩展，我们保持目标实体o不变，但扩展时间范围。这种编辑操作模拟了某些事实在未来并未发生改变的情况。例如，通过多次编辑（ME）后，模型中的美国领导人从2021年到2022年被编辑为拜登。然而，在实际情况中，从2022年到2023年，拜登仍然是美国领导人。因此，我们设计了扩展时间数据集并构造问题，以验证模型在这种情景下的知识编辑能力。
Extending Edit (EE) is an edit operation that extends the time scope of the current fact of the model. We traverse all chains of temporal facts and select the first fact in each chain as the extending object. By manually extending this object, we keep the target entity $o$ unchanged but extend the time scope. This edit operation simulates the situation where certain facts are not changed in the future. 
% For example, after passing multiple edits (ME), the US leader is edited to Biden from 2021 to 2022. However, in the real case, Biden remains the US leader from 2022 to 2023. Therefore, we design the extending time dataset and construct questions to validate the model's ability in this scenario.

% Examples of the three datasets, are shown in Table \ref{tab:dataset}, containing editing operations, and the corresponding questions and answers.

\begin{table*}[!t]
\centering
\resizebox{\textwidth}{!}{%
\begin{tabular}{ccccccccccccccc}
\toprule
\multirow{3}{*}{Method} & \multicolumn{5}{c}{\textsc{AToKe}-SE}          & \multicolumn{6}{c}{\textsc{AToKe}-ME}                    & \multicolumn{3}{c}{\textsc{AToKe}-EE} \\ \cmidrule(lr){2-6} \cmidrule(lr){7-12} \cmidrule(lr){13-15} 
 & \multicolumn{3}{c}{Current} & \multicolumn{2}{c}{Historical} & \multicolumn{3}{c}{Current} & \multicolumn{2}{c}{Historical} & Edited & \multicolumn{3}{c}{Current} \\ \cmidrule(lr){2-4} \cmidrule(lr){5-6} \cmidrule(lr){7-9} \cmidrule(lr){10-11} \cmidrule(lr){12-12} \cmidrule(lr){13-15}
                        & CES   & CES-P & CRS   & HES  & HRS  & CES   & CES-P & CRS   & HES   & HRS  & HES$^*$   & CES     & CES-P   & CRS    \\ \midrule
CFT                      & 5.73  & 5.69  & 5.34  & 0.06 & 0.02 & 1.11  & 1.18  & 1.22  & 0.03 & 0.00    & 0.01 & 3.41    & 2.91    & 3.15   \\
MEND                    & 80.47 & 40.56 & 32.46 & 1.73 & 0.68 & 71.83 & 27.96 & 23.67 & 0.40 & \textbf{2.10}  & 0.25  & 91.94   & 62.48   & 51.63  \\
ROME                    & \textbf{99.99} & \textbf{97.01} & \textbf{81.64} & \textbf{2.41} & \textbf{1.56} & \textbf{98.85} & \textbf{91.54} & \textbf{77.08} & 0.44  & 1.17 & 0.26  & \textbf{99.93}   & \textbf{98.70}    & \textbf{85.84}  \\
MEMIT                   & 99.66 & 92.23 & 75.31 & 2.22 & 1.21 & 98.42 & 91.06 & 66.48 & \textbf{0.48}  & 0.86 & \textbf{0.27}  & 99.92   & 95.82   & 72.76  \\ \bottomrule
\end{tabular}
}
\caption{Results of existing models on the benchmark \textsc{AToKe}-SE, \textsc{AToKe}-ME and \textsc{AToKe}-EE. ``Edited" means the score is computed when all the multiple edits are completed. The best results are highlighted in \textbf{BOLD}.}
\label{tab: baselineres}
\end{table*}

\subsection{Dataset Summary}

\subsubsection{Dataset format}
% 正如表一所示，
% 所有数据集，都还有 别名，和id
% SE 包含几个问题
% ME 包含几个问题。

% 三个数据集具有相似的数据格式，但它们在编辑次数和用于评估编辑方法的问题上有所不同。这些数据集都包含编辑集合（SE&ME&EE），问题（Q），问题的标准答案（A）和答案的别名集合。在这些数据集中，单次编辑和多次编辑都包含有关过去、现在和具体时间区间的问题，而扩展编辑仅包含有关当前和具体时间区间的问题。单次编辑和扩展编辑只有一组编辑集合、问题和答案，而多次编辑则包含多组。
The three datasets have a similar data format, but they differ in the number of edits and the questions used to evaluation. These datasets all contain edit sets (\textbf{SE\&ME\&EE}), questions ($\mathcal{Q}$), standardized answers to questions ($\mathcal{A}$), and answer alias sets. In these datasets, both single edit and multiple edits contain questions about historical and current knowledge, while extending edits contain only questions about the current knowledge. Single edit and extending edit have only one editing operation and corresponding questions for one $(s, r)$, while multiple edits contain multiple operations and questions. 
% Format examples for the three datasets can be found in the Appendix \ref{Appendix:Data Format}

\subsubsection{Data statistics}
% \begin{figure}
%   \centering
%   \begin{subfigure}[]{0.45\linewidth}
%     \centering
%     \includegraphics[width=\linewidth]{output_1.pdf}
%   \end{subfigure}%
%   \begin{subfigure}[]{0.51\linewidth}
%     \centering
%     \includegraphics[width=\linewidth]{output_2.pdf}
%   \end{subfigure}\\
%   \begin{subfigure}[]{0.5\linewidth}
%     \centering
%     \includegraphics[width=\linewidth]{output_3.pdf}
%   \end{subfigure}%
%   \begin{subfigure}[]{0.5\linewidth}
%     \centering
%     \includegraphics[width=\linewidth]{output_4.pdf}
%   \end{subfigure}
%   \caption{to be edited}
%   \label{fig:four_images}
% \end{figure}

% basic data中的关系数目的变化用文本描述。最后描述三个数据的个数都是多少个。
% 然后用四个图，来描述，两个描述。
% 第一个图，可以画一个加入之前的，再画一个加入之后的。
% =就是哪种柱状图，下面是不一样的颜色。
% 每个关系的编辑个数的图。每种关系作为横坐标
% SE和EE画在一张图。体现出从 gpt-j到编辑后的情况
% 考虑胰一下区间应该如何用图片描述。
% 实在不行用三张图/

% 在基础时序数据收集阶段，我们最初采样到了169,996个时序事实链。经过GPT-J的过滤，我们的时序链长度缩减为8,820，这也是三个数据集的最终长度。
% 对于时序事实链的长度分布，长度为2-5的事实链占整体数量的90%以上，极少数链的长度会超过10。此外，大多数关系的事实链长度为2，其中'<playsFor>'的关系的事实链最多且长度大于2的链占该关系整体的50%以上。更详细数据集统计和图像在附录中。
% 

In the base temporal data collection phase, we first sample 169,996 temporal fact chains. After filtering by GPT-J, there are 13 different relations left and the number of our temporal chains is reduced to 8,820, which is the final number of the three datasets. 
For the length distribution of temporal fact chains, fact chains of length 2-5 account for more than 90\%, with very few chains exceeding 10. In addition, the relationship "playsFor" has the largest number of fact chains. 
The time is measured in years. And for the time scope, before adding the future fake facts, the latest knowledge is obtained in 2022, and after that it is extended to 2028.
% More detailed statistics are in the Appendix \ref{Appendix:Data statistics}.

\subsection{Evaluation Metrics}
\label{sec:datametric}

% overal Question score
% Relative Time vs. Explicit Time 
% Previous relative Question score vs Current relative Question score
% HRS vs CRS
% Previous Explicit Question score vs Current Explicit Question score
% PES vs CES
% PCES
% 过去的基准主要关注知识模型中的知识本身，而且大多数知识编辑方法直接删除了过去的知识。然而，我们的基准侧重于评估模型对时间的感知能力，例如在学习新知识后是否保留旧知识。因此，在模型编辑过程中评估知识的时序性非常重要。我们主要通过对模型在不同时间的知识进行提问来进行评估。我们将时间相关的提问分为相对时间和绝对时间两个方面。绝对时间指的是针对模型提出具有明确时间范围的问题，例如："2003年至2007年，美国政府的领导人是谁？" 相对时间则不包含明确的时间范围，直接针对"当前"和"过去"进行提问，例如："现在美国政府的领导人是谁？" 在这基础上，相对时间和绝对时间又分为对过去和对现在的提问。因此，我们有以下四个指标来初步评估模型对知识的时序性。

%这一段可要可不要。 
% Previous benchmarks have focused primarily on the knowledge itself in the knowledge model, and most knowledge editing methods directly remove past knowledge. However, our benchmarks focus on assessing the model's ability to be time-aware, e.g., whether old knowledge is retained after learning new knowledge. Therefore, it is important to assess the temporality of knowledge during model editing. We primarily assess this by asking questions about the model's knowledge at different times.

As mentioned before, we categorize questions into relative time questions and explicit time questions, and both of them are about historical and current knowledge. 
Therefore, we have the following four indicators to assess the 
 performance of TKE:
1)\textbf{ Historical Relative time Question Score} (\textbf{HRS}), which is the accuracy of relative time questions about historical knowledge answered by the edited model. 
2) \textbf{Historical Explicit time Question Score} (\textbf{HES}), which is the accuracy of explicit time questions asked about historical knowledge.
3) \textbf{Current Relative time Question Score} (\textbf{CRS}), which is the accuracy of relative time questions about current knowledge. 
4) \textbf{Current Explicit time Question Score} (\textbf{CES}), which measures accuracy of explicit time questions about current knowledge.
In addition, the prompt we used as the optimization target when editing is the same as the questions in CES. Therefore, to make results more valid, we use paraphrases of the CES questions for further evaluation, denoted as 5) \textbf{Current Explicit time Paraphrase Question Score} (\textbf{CES-P}). 

% Finally, combining the above metrics, we give a composite score \textbf{S}. $$\textbf{S} = \frac{{\textbf{HRS} + \textbf{CRS} + \textbf{HES} + \frac{\textbf{CES}+\textbf{P-CES}}{2}}}{4}$$

% 描述，在mult下，和extend下特别的地方。
% 特别地，在多次编辑中，上述指标为模型多次编辑结束后的测量结果。HRS和CRS仅表示对最后一次编辑后的知识提问的结果，而HES和CES则是在编辑结束后对整个编辑过程中每个时间区间的知识提问结果进行平均计算的。
In particular, in {\dsme}, we treat each edit as a single edit and compute the score as in {\dsse}, and finally report the average of all edits. In addition, we calculate HES for all the historical facts at the end of multiple consecutive edits of the model to measure the overall editing performance (denoted as HES$^*$).
In {\dsee}, since there is no historical knowledge, we do not need to measure HRS and HES.
% , i.e., we do not need to measure HRS and HES.
% 我们的指标也有，和ES那个文章对应上。
% 上述指标均表示模型成功回答问题的正确率，即成功解码出答案。然而，有些研究将模型在新知识上的成功概率高于旧知识的概率视为编辑成功。基于这一观点，我们也进行了相应的测量，并在附录中提供了结果。
% All of the above metrics indicate the correctness of the model in successfully answering the question. However, some studies regard the probability of a model's success on correct object as higher than that on original object as editorial success. Based on this viewpoint, we also conduct the corresponding measurements. 
% The results of detailed formal definitions and another metric can be found in the Appendix \ref{Appendix:Evaluation Metrics}.

\section{Current Status on TKE}
In order to better assess the performance of existing methods for temporal knowledge editing, we conduct experiments with several popular editing methods and analyze the results.

\subsection{Experimental Setup}
\subsubsection{Knowledge Editing Methods} 
We evaluate the following state-of-the-art editing methods on \textsc{AToKe}.
1) \textbf{Constrained Fine-tuning (CFT)} \cite{zhu2020modifying} performs gradient descent on the target knowledge to minimize the loss and set a norm constraint on model weight changes.
2) \textbf{MEND} \cite{DBLP:conf/iclr/MitchellLBFM22} learns a hypernetwork to produce weight updates using a low-rank decomposition of the gradient obtained by standard fine-tuning.
3) \textbf{ROME} \cite{meng2022locating} firstly localizes the factual knowledge in the model by causal tracing, and then treats the MLP modules as key-value stores to insert new knowledge by making a rank-one change.
4) \textbf{MEMIT} \cite{meng2022massediting}, a successor to ROME, can insert lots of memories at once by modifying the MLP weights of a range of critical layers.

\begin{figure*}[]  
    \centering  
    \includegraphics[width=0.85\textwidth]{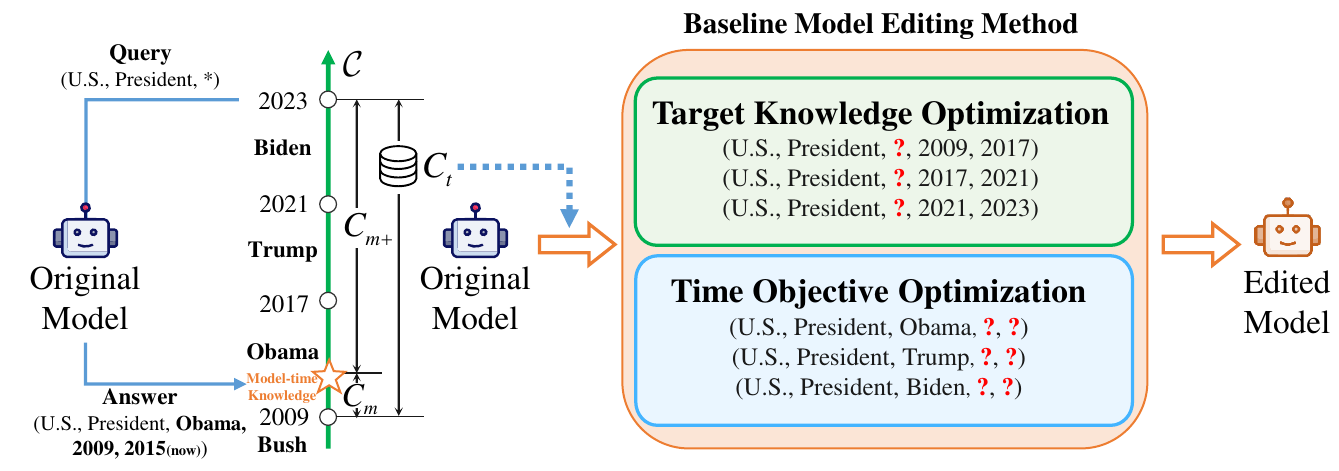} 
    \caption{Demonstration of the METO editing framework. First the model will be queried based on the current knowledge to get the knowledge under the model time ($C_m$). Then both historical and current knowledge are used as target knowledge for target knowledge optimization and time objective optimization with any model editing methods.}  
    \label{fig: meto}  
\end{figure*}

All of the above model editing methods make changes to the parameters of the model, which is the focus of our exploration. In addition, there are some methods based on external storage that do not edit the model. The performance of that class of methods depends on the knowledge classifiers or retrievers.
%, which is analyzed and explored in the Appendix \ref{sec:memory-base model}.

Given a knowledge editing operation that is expected to be learned $e = (s,r,o,t_s,t_u) \rightarrow (s,r,o^\star,t_s^\star,t_u^\star)$, we convert it to a cloze statement by natural language template $q_r(\cdot)$ which is used as the input to the above knowledge editing approaches.
Note that all the previous methods use only $q_r((s,r,o))$ as input; to ensure that the information is sufficient and adapted to our task, we have expanded these methods to include time for each input cloze statement.

\subsubsection{Language Model to be Edited}
Following setup of  previous work \cite{meng2022locating,meng2022massediting,zhong2023mquake}, we use GPT-J (6B) \cite{gpt-j} as the base LLM to be edited with above methods.

\subsection{Results and Analysis}
\label{sec: baselineres}
The performance of existing knowledge editing methods on our benchmarks is shown in Table \ref{tab: baselineres}.
For models, CFT is weaker and hence performs poorly. ROME performs slightly better than MEMIT, probably because we edit one piece of knowledge at a time and hence are better suited to ROME.
After comparison and analysis we can get the following main conclusions:
\paragraph{Remembering the new and forgetting the old} 
Except for CFT, edited models can remember new knowledge very well and do some generalization, but the performance on historical knowledge is disastrous. 
It is consistent with our expectations, as all existing model editing methods optimize the probability of current knowledge and ignore historical knowledge.
An example is shown in Figure \ref{fig: unedited}, where \textit{Donald Trump} is forgotten by the edited models.
\paragraph{Relative time questions are more difficult than explicit time questions}
We can observe that it is more difficult for the model to answer questions about a piece of knowledge without explicitly providing it with a specific time, using a relative time expression such as ``last one".
It may be because using relative time expression requires the model to reason about the order in which facts occur, whereas using explicit time expression simply requires the model to remember when facts occur.

\paragraph{The more edits, the worse the performance}
The average results on \textsc{AToKe}-ME are sightly worse than those on \textsc{AToKe}-SE, which may be because editing multiple times on the same fact causes a little confusion to the edited model. As we can see that HES score is only about 0.2\% after multiple edits, and all the knowledge injected before the last edit is invalidated.

\paragraph{Extending time scope is easier than inserting new object}
We can see that methods on \textsc{AToKe}-EE perform the best of three datasets, which is reasonable because the extending edit task does not change the current knowledge, but merely extends its time scope. The correct answer is not changed when the question is asked about the fact.

\section{METO: Multi-Editing with Time Objective}
As we can see from the previous experiments in Section \ref{sec: baselineres}, the existing methods perform excellently in memorizing new knowledge, but catastrophic forgetting of historical knowledge occurs after editing.
To alleviate this serious problem, we propose a simple editing framework METO (Multi-Editing with Time Objective) that can be applied to enhance existing editing methods easily.

\begin{table*}[]
\centering
\resizebox{\textwidth}{!}{%
\begin{tabular}{cccccccccccc}
\toprule
\multirow{3}{*}{Method} & \multicolumn{5}{c}{\textsc{AToKe}-SE}            & \multicolumn{6}{c}{\textsc{AToKe}-ME}                     \\  \cmidrule(lr){2-6} \cmidrule(lr){7-12}
 & \multicolumn{3}{c}{Current} & \multicolumn{2}{c}{Historical} & \multicolumn{3}{c}{Current} & \multicolumn{2}{c}{Historical} & Edited \\ \cmidrule(lr){2-4} \cmidrule(lr){5-6} \cmidrule(lr){7-9} \cmidrule(lr){10-11} \cmidrule(lr){12-12}
  & CES   & CES-P & CRS   & HES   & HRS   & CES   & CES-P & CRS   & HES   & HRS   & HES$^*$   \\ \midrule
CFT$^+$                      & 2.8\scriptsize{↓2.93}  & 2.62\scriptsize{↓3.07}  & 2.26\scriptsize{↓3.08}  & 3.38\scriptsize{↑3.32}  & 2.43\scriptsize{↑2.41}  & 1.27\scriptsize{↑0.16}  & 1.2\scriptsize{↑0.02}  & 1.04\scriptsize{↓0.18}  & 1.64\scriptsize{↑1.61}  & 1.23\scriptsize{↑1.23}   & 0.73\scriptsize{↑0.71}  \\
MEND$^+$                    & 83.26\scriptsize{↑2.79}  & 33.45\scriptsize{↓7.11} & 25.41\scriptsize{↓7.05} & 30.14\scriptsize{↑28.41}  & \textbf{30.17}\scriptsize{↑29.49}  & 70.52\scriptsize{↓1.31} & 28.41\scriptsize{↑0.45}  & 21.84\scriptsize{↓1.83} & 28.65\scriptsize{↑28.25}  & \textbf{30.83}\scriptsize{↑28.73}  & 21.83\scriptsize{↑21.58}  \\
ROME$^+$                    & \textbf{99.95}\scriptsize{↓0.04} & \textbf{93.78}\scriptsize{↓3.23} & \textbf{78.88}\scriptsize{↓2.76} & 20.25\scriptsize{↑17.84}  & 16.29\scriptsize{↑14.73}  & \textbf{99.93}\scriptsize{↑1.08}  & \textbf{90.97}\scriptsize{↓0.57} & \textbf{82.4}\scriptsize{↑5.32}   & 23.22\scriptsize{↑22.78}  & 18.18\scriptsize{↑17.01}  & 15.92\scriptsize{↑15.66}  \\
MEMIT$^+$                   & 86.4\scriptsize{↓13.26}  & 85.32\scriptsize{↓6.91}  & 74.07\scriptsize{↓1.24} & \textbf{30.31}\scriptsize{↑28.09}  & 24.32\scriptsize{↑23.11}  & 92.73\scriptsize{↓5.69} & 85.75\scriptsize{↓5.31} & 73.58\scriptsize{↑7.10}  & \textbf{36.2}\scriptsize{↑35.72}   & 26.04\scriptsize{↑25.18}  & \textbf{21.93}\scriptsize{↑21.66}  \\ \midrule
\end{tabular}
}
\caption{Results of enhanced models with METO (marked with ``$^+$") on the benchmark \textsc{AToKe}. Changes of the results are labeled on the side. The best results are highlighted in \textbf{BOLD}.}
\label{tab: metores}
\end{table*}

% img
\subsection{Methodology}
As shown in Figure \ref{fig: meto}, We firstly query the language model with the cloze statement of current knowledge to extract the knowledge under model time, and then edit the model using both of them with timestamps, and also use the model to make knowledge time predictions so that the model can reinforce both historical and new knowledge, and improve the awareness of time of knowledge.

\subsubsection{Model-time Knowledge Extraction}
For a chain of temporal facts $\mathcal{C}=\{(s,r,o_1,t_{s_1},t_{u_1}),...,(s,r,o_N,t_{s_N},t_{u_N})\}$, the object is constantly changing and the knowledge of the model about $(s,r,\cdot)$ is determined by the collection time of its training corpus. 
Therefore, in order to better preserve the historical knowledge inside the model, we need to extract the model-time knowledge about $(s,r,\cdot)$.
With the method in Section \ref{sec: modeltime}, we obtain the fact of what is happening under the model time which is noted as $C_m=\{(s,r,o_i, t_{s_i}, t_{u_i})\}$.
By comparing $\mathcal{C}$ and $C_m$, we can get the knowledge that needs to be newly captured by the model $C_{m+}=\{(s,r,o_{i+1},t_{s_{i+1}},t_{u_{i+1}}),...,(s,r,o_N,t_{s_N},t_{u_N})\}$. 

\subsubsection{Multi-editing on Both Historical and Current Knowledge}
Combining $C_m$ and $C_{m+}$, we obtain the set of all the target knowledge $C_t$ which is used as the target of editing.
Using the target knowledge $C_t$ obtained above as input, we can edit the language model with any model editing method, such as MEND, ROME, MEMIT, etc.
Note that in the specific implementation, to suit our task, we have also added timestamps in the cloze statement of original methods.

It is worth explaining that we only use $C_m$ which is occurring at the model time and do not use the full previous knowledge, $C_{m-}$, as the historical knowledge. It is because we believe that the model already knows that $C_{m-}$ is the history that has happened in the past which does not need to be changed. For $C_{m}$, on the other hand, the model needs to be made aware of the duration of this knowledge, which has ended and changed from the present tense to the past tense.

\subsubsection{Time Objective Optimization}
In order to further enhance the model's awareness of the time of knowledge, along with the multi-editing described above, we perform an additional task of optimizing the time objective of knowledge.
We also use the same editing method of existing model to ensure the generalization of our framework, except that the editing target is changed from the object to the corresponding time. As an example, given an input of ``Donald Trump is the President of the United States from", the model is then edited to optimize the probability of ``2017 to 2021" with ROME or other methods.

\subsection{Results on \textsc{AToKe}-SE and \textsc{AToKe}-ME}
Since existing methods perform relatively well on \textsc{AToKe}-EE and it does not involve the questioning of historical facts, we test with METO enhancement for editing on \textsc{AToKe}-SE and \textsc{AToKe}-ME (shown in Table \ref{tab: metores}).

We can find that maintaining little change in performance on current knowledge, our framework greatly improves existing editing methods' performance on historical knowledge.
Among them, ROME stills performs best in memorizing current knowledge. 
Surprisingly, MEND and MEMIT perform better on historical knowledge, which may be due to the fact that our framework prefers that editing methods can edit more than one piece of knowledge at a time, which in turn can preserve historical knowledge.

The editing experiments with our framework also further validate our two previous conclusions: 1) relative time questions are more difficult than explicit
time questions. 2) the more edits, the worse the performance.
It is worth noting that both phenomena have been mitigated to some extent.

It is promising that, as shown in Table \ref{tab: metores}, the HES$^*$ score on \textsc{AToKe}-ME improves from the original result of less than 0.3 to more than 15 (except for CFT). Such a significant improvement in this most difficult metric also shows that our framework is beneficial and the edited model remembers some of the previously injected knowledge.

Although there has been a substantial improvement in memorizing historical knowledge, it is still far from a satisfactory level, reflecting the difficulty of our proposed task of temporal knowledge editing, which still requires a concerted effort by the community.

\section{Related Work}
%\paragraph{Knowledge Editing}
The expanding parameter count of language models leads to higher retraining costs. And since the knowledge inside a model becomes progressively outdated, knowledge editing (KE), a convenient way to edit the knowledge, has received increasing attention and some methods have been proposed.
\citet{zhu2020modifying} propose the constrained finetuning approach on modified facts to solve the problem.
\citet{sotoudeh2019correcting} utilize symbolic representations and generalized RELU networks. \cite{article_relu} to correct model with small patches.
\citet{DBLP:conf/emnlp/CaoAT21} introduce the method which corrects knowledge and improves predictions using a hyper-network for targeted modifications.
\citet{dai-etal-2022-knowledge} explore knowledge neurons in models and attempt to leverage them to edit specific factual knowledge.
ROME \cite{meng2022locating} and MEMIT \cite{meng2022massediting} take the approach of locating and then editing.
\citet{DBLP:conf/iclr/MitchellLBFM22} a collection of small auxiliary editing networks that use a single desired input-output pair to make fast, local edits.
In addition, MEMIT-CSK \cite{gupta2023editing} is designed to edit commonsense knowledge in GPT.

Evaluation metrics on KE generally focus on whether the editing is successful and whether other unrelated knowledge is affected.
\citet{onoe2023lms} also evaluate abilities of updated LLM by making inferences based on injected facts.
Furthermore, \citet{zhong2023mquake} build multi-hop questions to assess edited models on related facts. And \citet{cohen2023evaluating} evaluate the ripple effects in the edited models.
However, no work has yet noted that pre-editing historical knowledge should also be preserved, which we argue is very important.

% \paragraph{Temporal Knowledge Graph}

\section{Conclusion}
In this paper, we systematically classify scenarios involving knowledge editing, identify the shortcomings leading to historical knowledge distortion in existing model editing methods. 
To facilitate a comprehensive evaluation of KE techniques, we introduce the task of temporal knowledge editing (TKE) and present a new benchmark named \textsc{AToKe}. We conduct experiments on \textsc{AToKe} and demonstrate that existing methods lead to a catastrophic forgetting of historical knowledge.
To bridging existing gaps, we present the METO editing framework, enhancing the efficacy of preceding approaches.
However, TKE still remains challenging and calls for more efforts in the community.

% \appendix
%\section{Reference Examples}
%\label{sec:reference_examples}

\section{Acknowledgments}
This work was supported by National Key R\&D Program of China (2021YFF0901502), National Science Foundation of China (No. 62161160339), State Key Laboratory of Media Convergence Production Technology and Systems and Key Laboratory of Science, Technology and Standard in Press Industry (Key Laboratory of Intelligent Press Media Technology). We appreciate the anonymous reviewers for their helpful comments. Xiaojun Wan is the corresponding author.

\bibliography{aaai24}

\end{document}